# Crypto Accounting Bench

Evaluating Frontier and Open-Weight Models on Crypto-Asset Accounting Tasks

**Kareem Khattab, Omar Khattab, Mohamed Ibrahem**
Entendre Finance

### Abstract

We introduce Crypto Accounting Bench (CAB), a benchmark for assessing whether frontier and open-weight language models can reconstruct the complete journal entry that an organization actually posted for a crypto-asset transaction. CAB contains 118 evaluation tasks drawn from 7 pseudonymized organizations. Each task combines transaction mechanics, asset quantities and base-currency values, wallet and legal-entity context, counterparty evidence, related transaction legs, recurrence, tax-lot evidence, and the organization's complete chart of accounts. The target is a balanced structured entry with every required account, side, amount, currency, and full-precision asset quantity. We evaluate 12 models spanning proprietary frontier systems and open-weight releases over 3 independent attempts per task, producing 4,248 trajectories. We report 3 metrics: Mean Score, Best@3, and Pass@3. Pass@3 is the fraction of tasks with at least 1 of 3 attempts that satisfies every rubric criterion and required gate. The leading model reaches 77.43% Mean Score, while the best Pass@3 is 56.78%. Deterministic diagnostics, read from each task's best of 3 attempts and macro-averaged across the 12 models, show higher base-amount agreement (97.8%) than deciding-account accuracy (56.3%). Together with the failure analysis, these results identify account selection and complete-entry composition as the main remaining challenges on CAB.

## 1 Introduction

Crypto-asset accounting requires more than reading a transaction hash or mapping an activity label to a generic account. The recorded entry depends on organization-specific wallet structures, legal entities, counterparties, internal conventions, account charts, linked transaction legs, and tax-lot evidence where assets are disposed. A model can identify the asset movement correctly and still post the wrong income, expense, clearing, intercompany, payable, or gain/loss account. Accounting research documents this variability directly: practice varied widely across U.S. public firms holding crypto assets from 2013 to 2022, which motivated the dedicated standard ASU 2023-08 [1]. CAB therefore targets the treatment an organization recorded, not a normative ideal.

Existing finance benchmarks test financial knowledge, auditing, synthetic workflows, or broad professional deliverables. CAB instead asks a narrow question with a checkable answer: given the same transaction evidence available to the accounting workflow, can a model reproduce the full entry that was posted? The answer is a balanced, organization-specific journal entry, not a narrative recommendation, and its lines can be compared directly with the entry the organization recorded.

### Contributions

CAB contributes complete-entry reconstruction as an evaluation task over real crypto-asset accounting evidence, and a posted-entry answer key that is withheld from model inputs and paired with a pseudonymized task preserving the evidence needed to infer it. It also contributes an evaluation that combines frozen task-local rubrics, LLM-assisted criterion grading with locally summed weights, and independent deterministic diagnostics; and results for 12 models over 3 attempts each, whose capability slices place the failures in account selection and complete-entry composition.

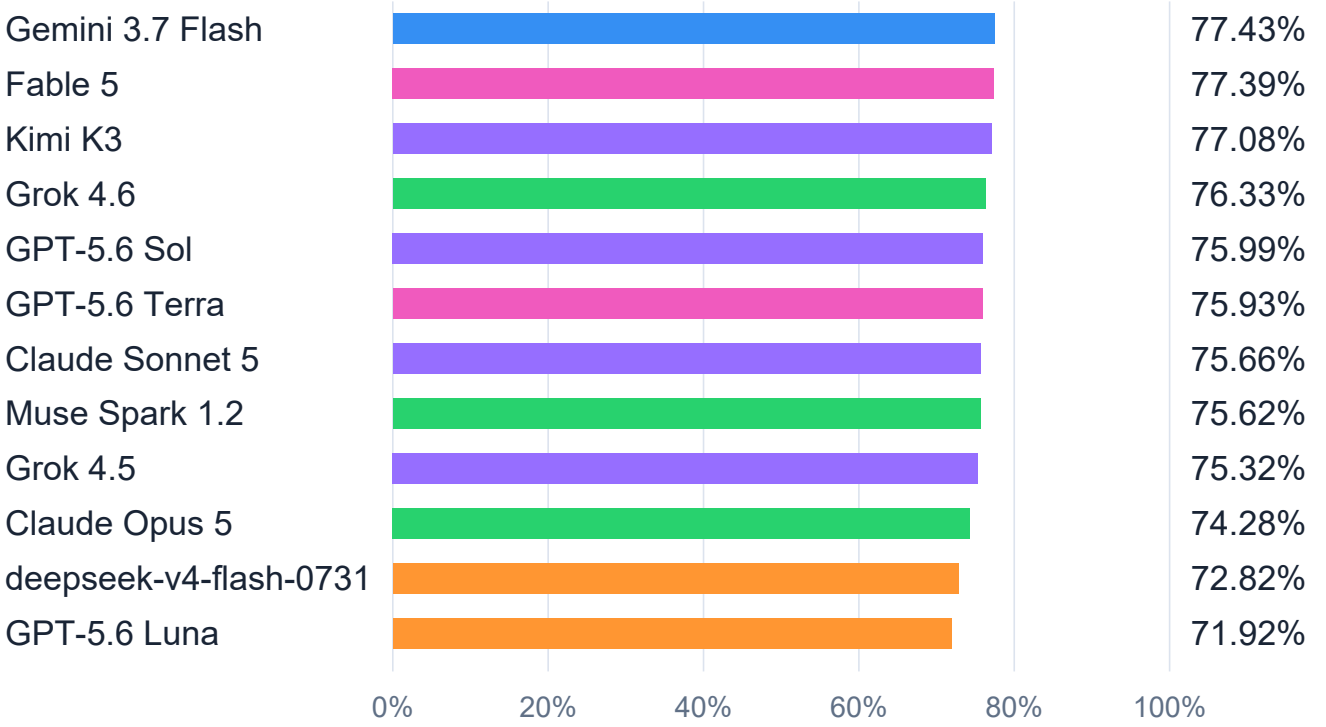

*Figure 1. Mean Score across the 118 evaluation tasks, averaged over all 3 independent attempts per task. The top of the leaderboard is tightly clustered, so values are printed to 2 decimal places.*

The score spread is narrow: Gemini 3.7 Flash leads at 77.43%, while GPT-5.6 Luna records 71.92%. No model comes close to reliable Pass@3 success, which requires every rubric criterion and every required gate on 1 attempt (Section 6).

### Evaluation question

Can a frontier model transform heterogeneous transaction evidence into the exact balanced entry used by the organization without access to the answer key?

## 2 Related Work

**Professional-task benchmarks.** Evaluation has moved from exam-style questions toward economically meaningful deliverables. GDPval [2] spans 44 occupations across 9 GDP-contributing sectors and grades professional outputs produced by industry practitioners. It motivates CAB's emphasis on complete outcomes and repeated attempts, while CAB differs in demanding 1 structured accounting artifact checked against ground truth taken from a real ledger.

**Finance and accounting workflow benchmarks.** FinMaster [3] combines a simulator with 183 tasks covering financial literacy, accounting, auditing, and consulting over generated company data. AuditBench [4] pairs real financial tables with synthesized transactions and evaluates error detection, explanation, and mapping to accounting standards through a 5-stage framework. Finch [5] evaluates spreadsheet-centric enterprise finance workflows built from authentic enterprise artifacts, with 384 tasks across 172 multi-step workflows. These measure broad workflow competence over many artifacts; CAB measures 1 decision at transaction level.

**Closest accounting benchmarks.** FinBalance [6] reconciles source-document bundles into cited journal entries and a balance sheet, validated by deterministic ledger replay. Its documents are composed by a deterministic generator from human-authored scenarios. APEX-Accounting [7] evaluates 160 month-end close tasks across 10 synthetically generated company worlds against rubrics of binary criteria. CAB differs from both in what supplies the answer key and in how narrow the unit of work is. Each task is 1 crypto-asset transaction, and the target is the complete entry that 7 real organizations posted in their own ledgers. The model selects from each organization's chart of accounts, given wallet, legal-entity, counterparty, chain, related-leg, recurrence, and tax-lot evidence.

### Positioning

CAB's target is descriptive and organization-specific: the complete entry that exists in the source ledger. This design assesses whether a model can combine the supplied evidence, select the appropriate organization-specific accounts, preserve numeric precision, and satisfy double-entry constraints.

# 3 Benchmark Design

CAB is built as a one-way pipeline: recorded entries and their source evidence go in, and only a pseudonymized, model-visible task comes out.

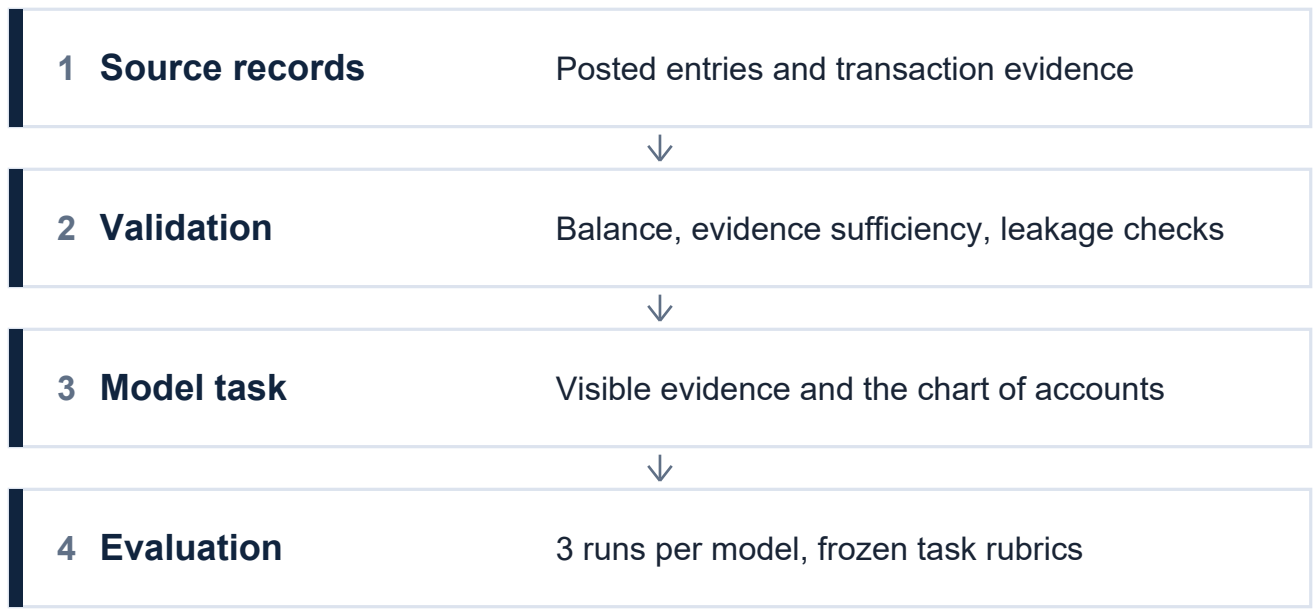

*Figure 2. CAB production and evaluation pipeline, from source records and validation to model-visible tasks and evaluation.*

## 3.1 Task construction

Each task starts from a posted journal entry joined to its source transaction and available accounting context. The task renderer projects factual evidence into a model-visible schema. Every task in the evaluation set meets the same construction and validation requirements: a non-degenerate balanced entry, a complete chart of accounts, reproducible monetary amounts, reachable account choices, and enough visible evidence to infer every non-zero reference line. The reference entry is the journal entry the organization recorded, and it is the evaluation target.

## 3.2 Information available to the model

The model receives: transaction type and date; movement direction; asset, chain, quantities, values, and unit price; source wallet or exchange account and legal entity; counterparty and external labels; contract and function metadata where available; same-hash related transactions; prior recurrence; created or relieved tax lots; and the organization's complete chart of accounts. We exclude the reference entry itself, labels identifying the correct accounts, private source identifiers, and authored solution notes. Transaction values, tax-lot evidence, and the organization's complete chart of accounts remain visible as task inputs.

# 4 Dataset

The evaluation set contains 118 tasks from 7 pseudonymized organizations, spanning 21 assets, 10 chains plus an exchange-only context, 74 deciding accounts, 2 base currencies, and 8 recorded source-system labels. Appendix B reports the evidence available across the set.

## 4.1 Ground truth

The target for every task is a structured record containing all posted lines, account names, Debit/Credit sides, base-currency amounts, currency, and the transaction's full-precision asset quantity. The evaluation set contains 246 non-zero reference lines. All reference entries balance at accounting precision and all graded amounts are supported by a visible evidence family or by double-entry closure from already measured lines.

## 4.2 Accounting situations covered

Grouped by the treatment their reference entry records, the set covers transfers, income and expense events, intercompany transfers among the 20 legal entities in the corpus, swaps, fee events, and disposals that relieve a tax lot and book the resulting realized gain or loss.

The asset mix varies as well. 67 tasks move a dollar-pegged stablecoin, 6 of them a vault-bridge token representing an asset custodied on another chain, and 3 move a share in a tokenized treasury fund. Another 3 involve wrapped or liquid-staked positions, and 14 sit on Bitcoin or the Babylon staking chain. A card issuer contributes customer card-settlement receivables, collateral funding, a loan drawdown, and stablecoin redemption and clearing decisions.

23 tasks are exchange-sourced rather than on-chain. They record realized trading gains and losses, funding income received, exchange trading fees, and 1 loan drawdown. No single trading account absorbs them. The correct account depends on the exchange, on the internal trading strategy the flow belongs to, and on whether value entered or left the account. An otherwise identical stablecoin inflow can therefore belong to one strategy's gain account or to another's. The evidence carries the cash and token movements these accounts generate rather than the underlying positions, so CAB does not test derivative position accounting.

## Why reconstruction is hard

The search space is wide before any judgment is applied: a task ships a median of 569 chart accounts, and the 118 recorded entries between them post to only 141 distinct accounts. Nearly every account a model can see is the wrong answer for any given task.

Reconstruction rarely follows from the transaction alone. Selecting the deciding account can require the wallet's own name and custody role, the legal entity on each side, whether the counterparty is another wallet of the same group, the exchange and trading strategy a flow belongs to, token contract metadata, and the prior posting history of similar recurring transactions. The corpus supplies this evidence unevenly. In 64 of the 118 tasks, the evidence identifies a legal entity on the other side of the transaction. The model must compare it with the source wallet's legal entity: the entities match in 42 tasks and differ in 22 intercompany tasks. Where a disposal relieves a lot, the entry must also carry the relieved carrying basis and the realized gain or loss. Direction and amount are the parts the evaluated models usually get right. The distance between 97.8% base-amount agreement and 56.3% deciding-account accuracy, both macro-averaged best-of-3 diagnostics across the 12 models (Section 7.1), highlights the accounting judgment challenge this benchmark targets.

## 4.3 Task anatomy

A task asks the model to infer 1 complete entry from the evidence taken together. The contract requires valid JSON and forbids invented accounts. A simplified illustration is shown below; names and values are illustrative.

```
{
  "journalEntry": {"lines": [
    {"ledgerAccountName": "11157: TokenA Wallet",
     "drCr": "Debit", "amountBase": "0.24",
     "currency": "USD"},
    {"ledgerAccountName": "71008: Dividend Income",
     "drCr": "Credit", "amountBase": "0.24",
     "currency": "USD"}
  ]},
  "assetQuantity": "..."
}
```

## 4.4 Privacy

The evaluated corpus uses pseudonymized identifying names and keeps private database identifiers outside model inputs. The public Hugging Face release is a further transformed derivative of the same 118 underlying cases rather than the exact inputs used in the reported evaluation. It replaces identifying names, account codes, blockchain identifiers, and asset and chain labels; rescales monetary values and asset quantities; and shifts dates while reducing timestamp precision. Reference answers are transformed consistently, with gain/loss amounts reconciled to the transformed evidence. Generic accounting terminology and some public categorical labels remain. The answer key is stored apart from the model-visible corpus. Source-ledger provenance, real organization identities and the transformation mappings are not released, and these measures do not establish complete anonymity.

# 5 Experimental Setup

## 5.1 Models and harness

We evaluate 12 models, spanning proprietary frontier systems and open-weight releases served through hosted inference. Each model receives the same rendered task prompt, the same system contract, and 1 exposed tool type: a bash command runner. A model may call it repeatedly across up to 12 agent turns. Each attempt runs in a fresh temporary working directory, and no reference answer or benchmark-private artifact is copied into it; evaluation loads the reference answer separately, after candidate inference. Each command carries a 60-second timeout, and all production pipelines set a 32K output-token ceiling. Runs use the highest reasoning setting each pipeline exposes.

That working directory is workspace isolation rather than an operating-system filesystem or network sandbox, and provider API timeouts and reasoning controls are pipeline-specific. Each model executes every task independently 3 times, so each model runs 354 task attempts and the evaluation records 4,248 trajectories in total. 1 trajectory is 1 task attempt. Appendix A reports the per-model reasoning setting, the settings shared across the 12 pipelines, and the judge configuration.

### 5.2 LLM-assisted rubric evaluation

Scoring and validation use 4 complementary layers: **(1) A frozen rubric** gives each task 6 binary criteria, fixed by its accounting family and derived from its ground-truth entry; the criteria and their weights belong to the task definition, and no model contributes to them. **(2) Criterion grading** has an LLM judge return 1 PASS/FAIL verdict per criterion, reading the task evidence, the criterion, the candidate's final structured answer, and the reference answer. **(3) Local summation** sums the weighted score from those verdicts and the frozen weights. **(4) Deterministic diagnostics** compute balance, line matching, account, side, amount, currency, quantity, and normalized complete-entry equality from the answer itself, independently of the judge, and supply the structural pass gates.

The judge grades only the supplied material and holds no authority over weights, thresholds, or gates. 1 fixed judge configuration served every reported evaluation, and Appendix A states it. Each layer is recorded per attempt, so every reported score follows from the stored artifacts alone.

| Criterion family | Weight |
|---|---|
| Accounting treatment | 25% |
| Primary / deciding account | 25% |
| Counter account | 15% |
| Debit / Credit treatment | 15% |
| Accounting amount(s) | 15% |
| Complete required entry | 5% |

*Table 1. Weights for the transfer, fee and income/expense families, 76 of the 118 tasks; the treatment label varies by family. Intercompany, swap and realized gain/loss tasks use their own frozen 6-criterion rubrics. Every family totals 100%.*

### 5.3 Metrics

CAB reports 3 metrics. Write the weighted rubric score of attempt $a$ on task $t$ as $s(t, a)$, a number in [0, 1], and let each task have 3 attempts.

**Mean Score** is the average of $s$ over every expected attempt: the average partial-credit score of 1 try. **Best@3** takes each task's strongest attempt, $\max_a s(t, a)$, and averages that over tasks; it is an oracle-style capability ceiling rather than a reliability figure. **Pass@3** is the fraction of tasks for which some attempt earns the full rubric score *and* clears every required gate. Writing $g(t, a)$ for the conjunction of the required gates on that attempt:

$$\text{Pass@3} = |\{\, t : s(t, a) = 1.00 \text{ and } g(t, a) \text{ for some attempt } a \,\}| \div |T|$$

Both conditions must hold on the **same** attempt. Partial credit never qualifies: 0.99 is not a pass, and neither a perfect rubric score that trips a required gate nor clean gates without a perfect score is a pass. Pass@3 therefore measures full rubric-and-gate success across 3 attempts. It does not require normalized complete-entry equality, which the deterministic checker evaluates separately and which is not among the gates. We also report component diagnostics.

### Critical accounting gates

The rubric measures accounting correctness. Alongside it, 7 required gates run on every attempt. 6 are deterministic. 5 of those are computed from the candidate answer itself: parseable output, balance, the required journal structure, quantity consistency, and wallet custody correctness. The 6th verifies that the task's reference entry is gradeable. The 7th, material accounting correctness, is computed from the same frozen criterion verdicts the rubric uses.

The rubric and the gates ask different questions. The rubric asks whether the accounting is right; the gates additionally require the journal structure the organization recorded and the account names it used. An attempt can therefore satisfy every accounting criterion and still be rejected, for example by combining 2 recorded postings into 1 posting of the same net amount, or by writing an account name the reference does not use. The gates impose structural requirements in addition to rubric correctness. Appendix A gives the gate definitions, how often the 2 disagree, and the calibration used to check them.

### 5.4 Run identity

All reported evaluations use the same task rubrics, criteria, weights, judge configuration and deterministic diagnostics. 1 grading contract therefore covers all 4,248 trajectories, and every reported attempt is scored under it.

Every artifact records the benchmark identity, the dataset, the system-prompt, task-prompt and task-local rubric hashes, the evaluator and rubric identities, the judge configuration and reasoning setting, and the hash of the candidate response it grades. An evaluation and the attempt it scores can therefore be matched exactly, and every aggregate in this paper follows from the stored per-attempt evaluations.

## 6 Main Results

Table 2 reports the leaderboard, ordered by Mean Score, the primary metric. No model leads on all 3. Gemini 3.7 Flash has the highest Mean Score at 77.43%, while Kimi K3 takes both Best@3 at 82.37% and Pass@3 at 56.78% from 3rd place on Mean Score.

| Model | Mean Score | Best@3 | Pass@3 |
|---|---|---|---|
| Gemini 3.7 Flash | 77.43% | 81.61% | 52.54% |
| Fable 5 | 77.39% | 81.74% | 54.24% |
| Kimi K3 | 77.08% | 82.37% | 56.78% |
| Grok 4.6 | 76.33% | 79.45% | 50.00% |
| GPT-5.6 Sol | 75.99% | 81.36% | 50.00% |
| GPT-5.6 Terra | 75.93% | 81.53% | 50.85% |
| Claude Sonnet 5 | 75.66% | 80.51% | 47.46% |
| Muse Spark 1.2 | 75.62% | 80.76% | 47.46% |
| Grok 4.5 | 75.32% | 80.25% | 48.31% |
| Claude Opus 5 | 74.28% | 79.11% | 45.76% |
| deepseek-v4-flash-0731 | 72.82% | 79.45% | 46.61% |
| GPT-5.6 Luna | 71.92% | 78.77% | 47.46% |

*Table 2. Crypto Accounting Bench leaderboard over 118 tasks and 3 attempts per model, ordered by Mean Score, which averages all 3 attempts. Pass@3 is defined in Section 5.3.*

### Interpretation

**Ranking compression.** The top 9 models span 2.10 percentage points on Mean Score. We report no confidence intervals, so the order inside that band is not a settled ranking. What the spacing does show is that these systems reach similar average partial credit even where their full-rubric success rates diverge.

**The partial-credit gap.** Every model exceeds 71% Mean Score, yet the best Pass@3 is 56.78%. Averaged over the 12 models, 50.21% of tasks yield no passing attempt out of 3. Models routinely recover the amount, the direction and 1 side of an entry, then misclassify the deciding accounting line.

### 6.1 Full-rubric success across models

Pass@3 ranges from 45.76% to 56.78% across the 12 candidates (Figure 3), against Mean Scores of 71.92% to 77.43%. Models earn substantial partial credit, but many tasks still have no fully passing attempt across 3 tries.

### 6.2 Rubric quality against completion

Figure 4 plots average rubric quality against Pass@3 success. Across the 12 models, Pass@3 correlates with Mean Score (Pearson r = 0.74) and with Best@3 (r = 0.82). Both correlations are over the 12 models, n = 12, rather than over individual attempts. 6 candidates reach the upper-right quadrant, and the best Pass@3 among them is 56.78%.

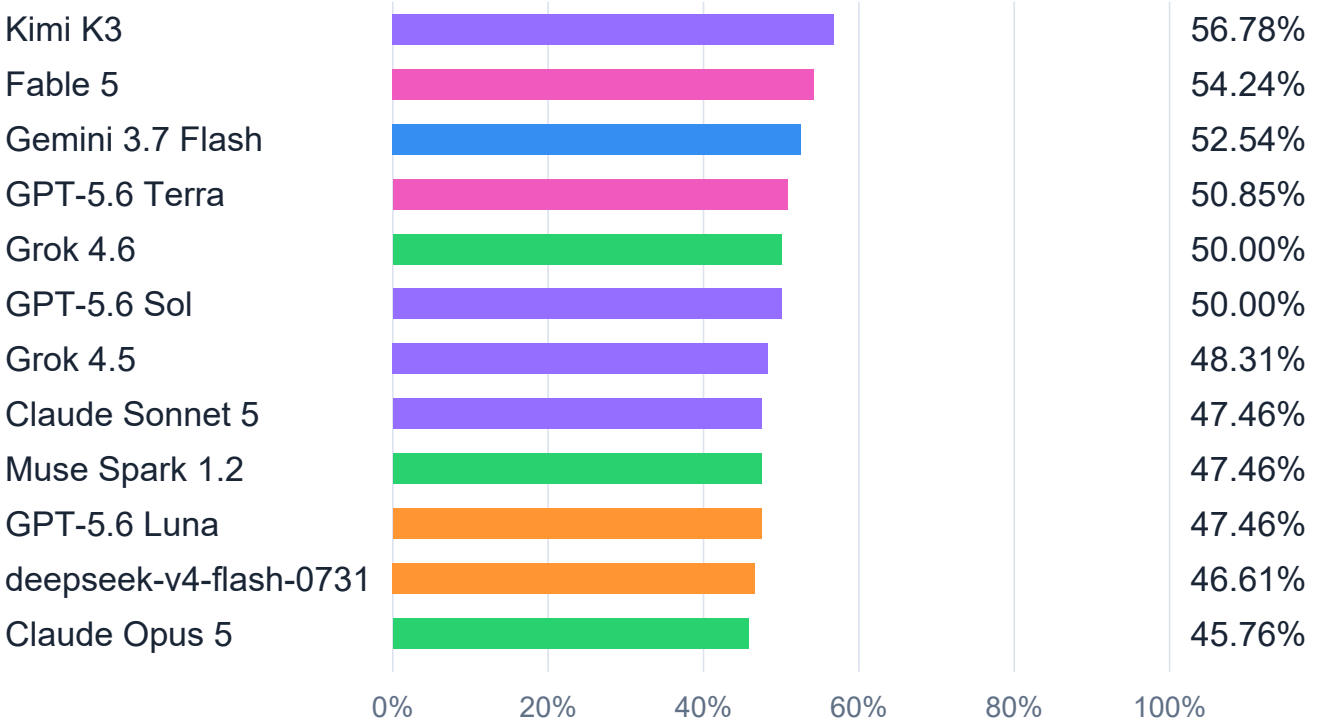

*Figure 3. Pass@3 by model, ordered by Pass@3, highest first; ties are ordered by Mean Score.*

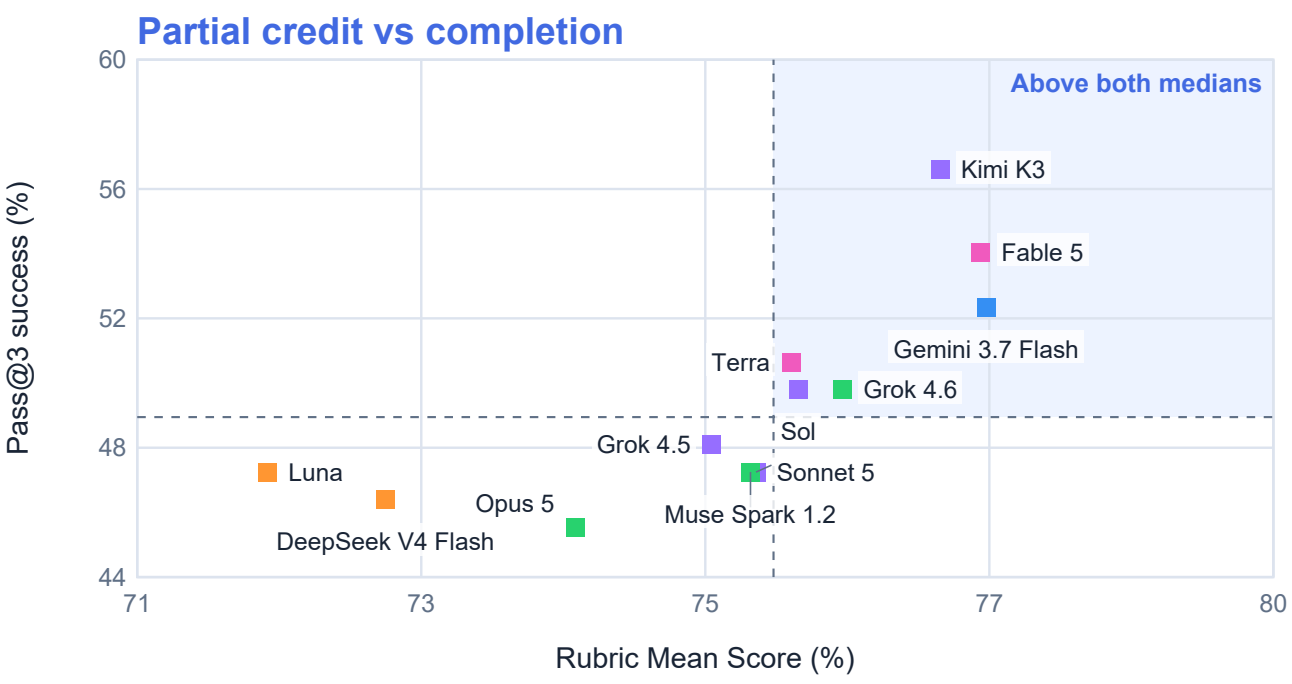

*Figure 4. Average rubric score against Pass@3 success, 1 marker per model, coloured as in Figure 1. Dashed lines mark the median on each axis; the shaded quadrant contains models above both medians.*

### 6.3 Score against inference cost

Average candidate inference cost per task attempt ranges from $0.019 to $0.454 across the 12 runs, against 5.51 points of Mean Score. We observe little association between cost and Mean Score in this evaluation (Pearson r = 0.26, n = 12). The highest average cost per task attempt does not lead on Mean Score. Fable 5 costs $0.454 per attempt and scores 77.39%, while Gemini 3.7 Flash leads at 77.43% for 52% of that cost. Kimi K3 takes both Best@3 and Pass@3 at $0.101 per task attempt. 3 models score above the median for less than the median average cost per task attempt.

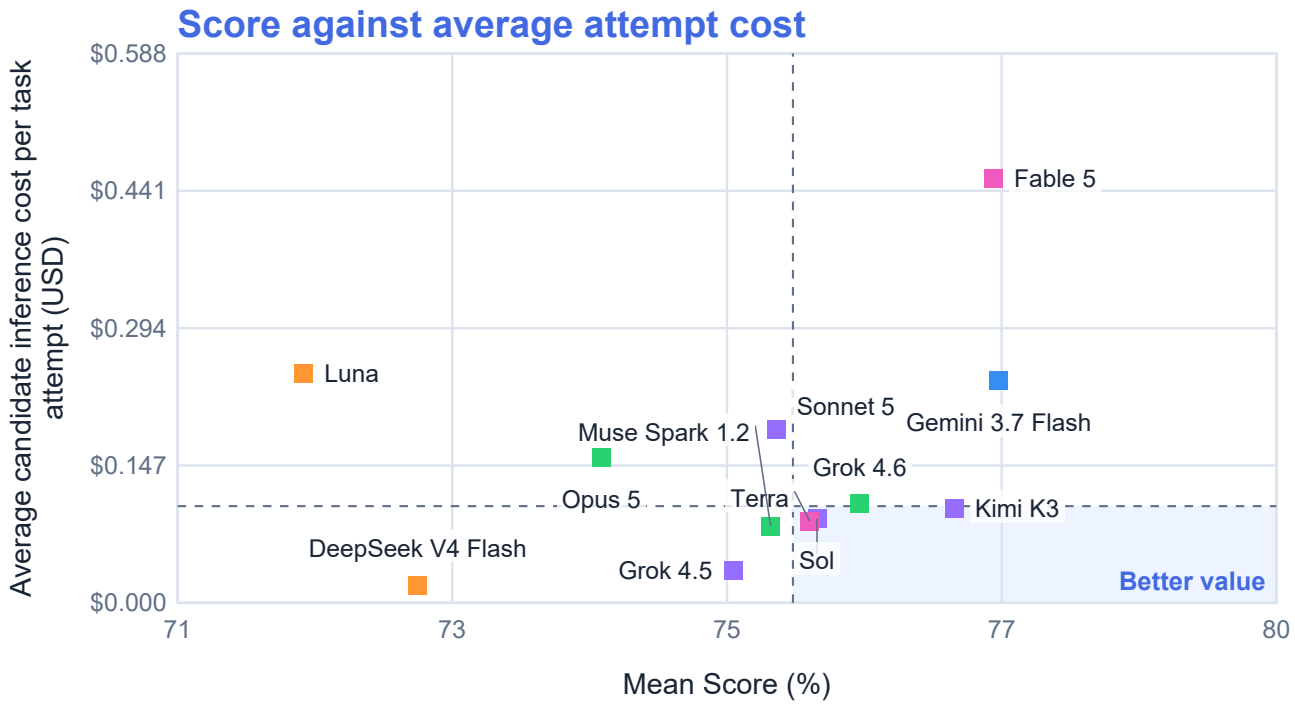

*Figure 5. Mean Score against average candidate inference cost per task attempt. Each marker represents 1 model; dashed lines mark the median on each axis.*

## 7 Capability Analysis

### 7.1 Where the errors concentrate

The component figures are deterministic diagnostics. For each model and task, we select the attempt with the highest rubric score, using the deterministic tie-breakers for ties, and compute all component diagnostics from that same answer. We average over tasks, then equally across the 12 models. Because the selection is on rubric score rather than on the component being reported, no diagnostic is maximized independently, and these are not single-attempt accuracies. On that basis asset quantity accuracy reaches 100.0%, accounting direction 98.1% and base-amount agreement 97.8%, while deciding-account accuracy is 56.3%. The 41.5-point gap places the errors in treatment and account selection rather than in arithmetic.

The 6 diagnostics measure different things. Asset quantity checks whether the response reproduces the transaction's asset quantity at the supplied precision, which is 1 value per transaction rather than a journal-line amount. Accounting direction measures whether assigned candidate and reference lines fall on the same Debit/Credit side, and account identity is not required for agreement. Base amount measures whether those lines carry the reference base-currency amounts. Account agreement is the share of reference lines assigned a candidate line with the correct account, so a partly correct entry earns partial credit. Line match F1 balances precision and recall over complete line matches, which require agreement on account, side, base-currency amount and currency. Deciding account records whether the response recovers the reference-marked account carrying the task's central accounting decision.

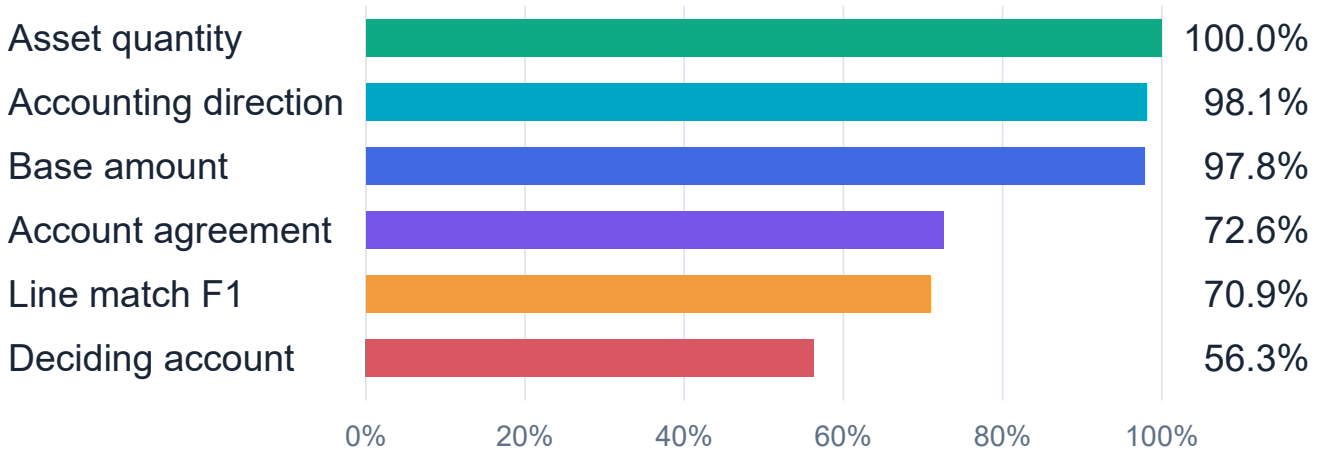

*Figure 6. Macro-average best-of-3 deterministic diagnostics across the 12 models. Each diagnostic is read from the same selected attempt per task.*

### 7.2 Rubric families

The 6 rubric families partition the 118 tasks, so Figure 7 covers the whole evaluation set. Rubric family, transaction label and derived entity relationship are 3 separate classifications. The corpus carries 22 transactions labelled INTERCOMPANY TRANSFER and 22 tasks with a derived intercompany entity relationship, and the 2 sets are not identical. Figure 7 groups tasks by grading rubric, where 21 take the INTERCOMPANY rubric; the remaining labelled transaction books a realized gain, so the realized gain/loss rubric applies. Fee and swap tasks reach the highest macro Mean Scores, at 87.5% and 84.5%, and income/expense and intercompany tasks the lowest, at 67.5% and 68.1%. The ordering tracks how much of the entry the transaction itself fixes. A fee posts to a fee account; an income or intercompany line has to be chosen from the organization's own chart, against evidence about the counterparty and the entities on each side. The realized gain/loss family sits mid-range at 76.5%. Given its sample size of 5 tasks, this family-level mean should be interpreted cautiously.

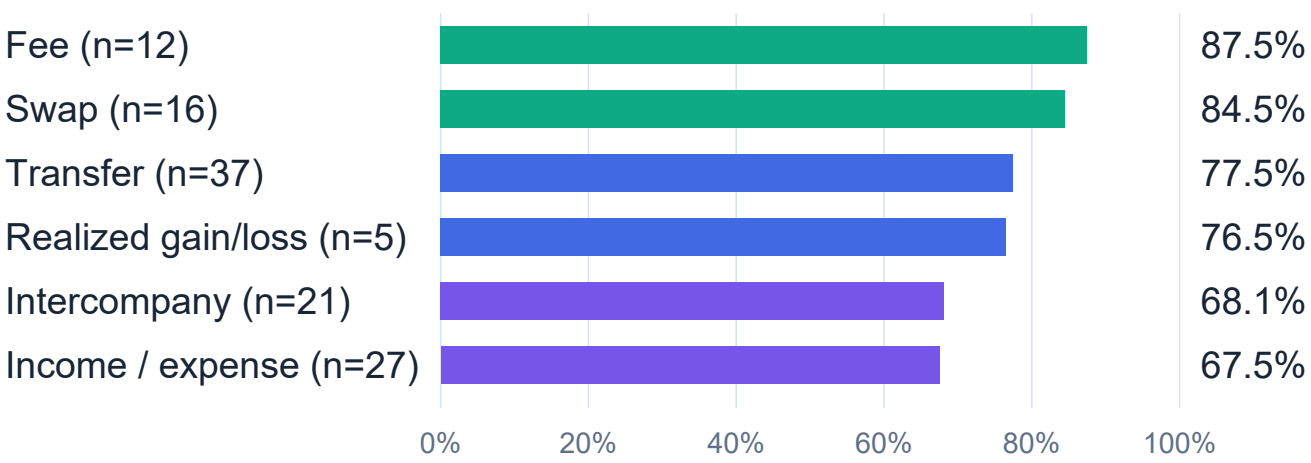

*Figure 7. Mean Score by rubric family, macro-averaged across models. Families group tasks by the rubric used to grade them and partition the set: the sample sizes in the labels total 118.*

## 8 Failure Analysis

Incomplete entries and misposted accounts are the most frequent signals, each appearing 2,328 times. The Debit/Credit criterion fails 2,243 times, unsupported postings appear 2,214 times, and the deciding account is wrong 1,997 times. Incorrect treatment accounts for 1,073, of which 454 are intercompany. An incorrect amount occurs only 198 times across 4,248 trajectories. Entry composition and account choice dominate numeric error, as the component diagnostics also show.

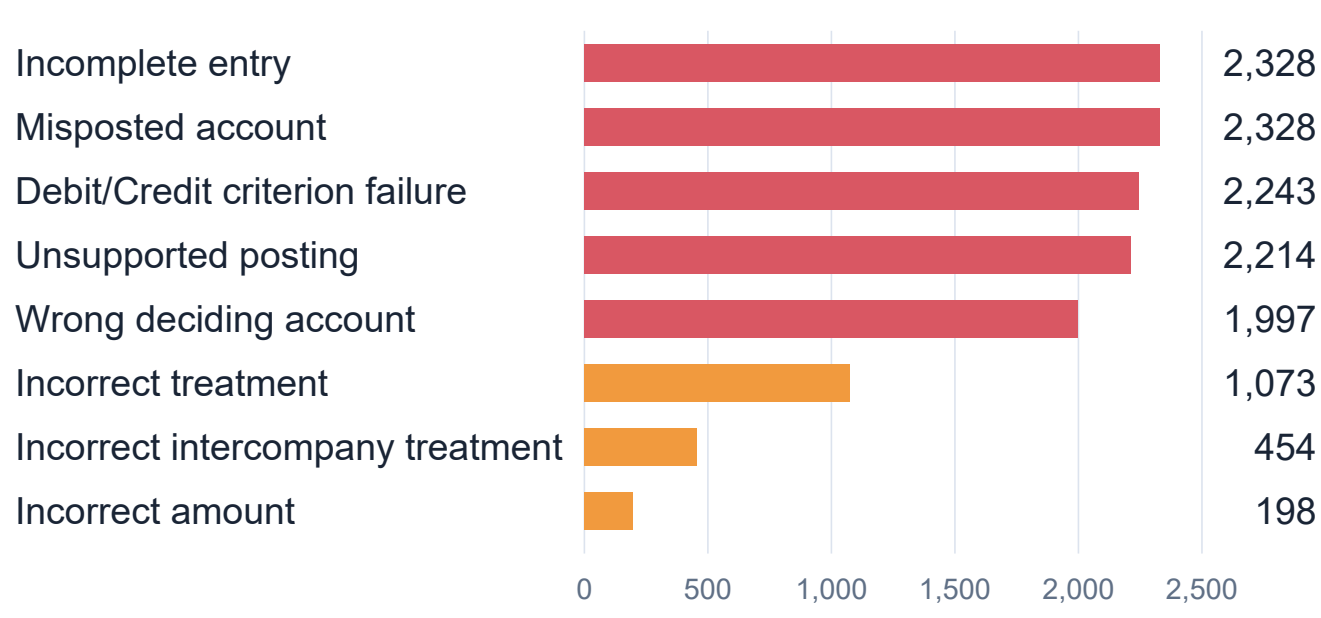

*Figure 8. Most common deterministic failure signals across 4,248 trajectories. Signals are non-exclusive and counted at most once per trajectory.*

### 8.1 Unmatched postings

The 2 leading signals describe 1 error from 2 angles. A line posted to an account the reference does not use is itself a misposted account, and it leaves the reference's own line without a counterpart, which registers as an incomplete entry. That is why the 2 totals are equal. On 2,141 of those trajectories the candidate carries exactly as many lines as the reference, so the dominant failure is substituting 1 account for another rather than stopping the entry short.

### 8.2 Correct family, wrong account

Models more often miss the account than the category: the deciding account is wrong 1,997 times against 1,073 incorrect treatments. The recurring pattern is a generic account in place of the organization-specific deciding account supported by wallet names, entity relations, contract metadata, or historical recurrence, which earns treatment-level partial credit while failing the primary-account and complete-entry criteria.

### 8.3 Intercompany direction

Intercompany tasks require both entity comparison and direction-sensitive account choice. An inflow may imply a payable while an outflow implies a receivable, but only after establishing that both wallets belong to different legal entities. The direction cannot be read off the movement alone.

### 8.4 Format compliance

Invalid responses are rare, and quantity and currency failures are similarly rare. The models generally follow the JSON contract and produce balanced candidates, and the dominant errors occur inside otherwise valid entries. On this evaluation set the models are separated by accounting judgment rather than by format compliance.

### Key finding

The 12 models evaluated here usually move the correct value in the correct direction, but none of them reliably identifies every organization-specific account or composes the complete journal entry.

## 9 Limitations and Responsible Use

CAB evaluates 1 operation: reconstruction of a posted crypto-asset journal entry. It does not evaluate month-end close, tax preparation, audit, consolidation, or external reporting. The benchmark is descriptive: it measures agreement with the organization's recorded entry and does not certify that the entry is the only normatively acceptable accounting policy. Results characterize performance on the CAB evaluation population described in Section 4.

LLM-assisted criterion grading introduces residual judgment uncertainty even when the ground truth and criterion are provided. Fixed weights, local summation, deterministic diagnostics, and structural gates provide consistency, traceability, and structural checks, but they do not independently validate the judge's accounting decisions. Pass@3, defined in Section 5.3, is the most demanding of the 3 reported metrics because it accepts no partial rubric credit.

The tasks and reference answers are derived from journal entries the organizations recorded in their own accounting systems. Each reference answer is the entry the organization posted, and CAB evaluates whether a model can reconstruct that entry from the supplied transaction evidence. The rubric criteria are authored from accounting practice rather than induced from model behaviour.

That provenance fixes the target, not its authority. It is not an independent audit of the entry, and it says nothing about the reliability of the automated grading described above, which remains a separate source of uncertainty.

### Responsible deployment

A benchmark pass is not authorization to post autonomously to production books. Real deployment should retain human review, access controls, provenance, reconciliation, and rollback. CAB is intended to diagnose model capability and guide system design, not to remove professional accountability.

## 10 Conclusion

Crypto Accounting Bench evaluates whether frontier models can reconstruct complete organization-specific journal entries from real-world crypto-asset transaction evidence.

Across 4,248 trajectories the best Mean Score is 77.43% and the best Pass@3 is 56.78%. Output syntax, balance, quantity and amounts are rarely where these models fail: on best-of-3 diagnostics asset quantity accuracy is 100.0% and base-amount agreement 97.8%. The errors that remain sit in accounting judgment, in choosing the deciding account and composing every line the entry needs.

Every model family evaluated shows the same pattern, with Pass@3 between 45.76% and 56.78%, which is consistent with a shared limitation rather than a weakness of 1 pipeline. Progress will require better evidence synthesis, organization-specific account selection, tax-lot reasoning, and explicit verification that every required line is present before submission.

### Data and evaluation artifacts

The public Crypto Accounting Bench educational dataset is available on Hugging Face and contains transformed counterparts of the same 118 underlying cases used in the reported evaluation, together with their schemas, manifest and reference answers, so this manuscript and the public release describe 1 benchmark population; Section 4.4 states what the transformation changes. Reference answers are excluded from model inputs and from the candidate workspace, and evaluation loads them separately. The open-source evaluation toolkit is on GitHub, covering the runner, model pipeline templates, deterministic scorer, validation suite, report generator and leaderboard generator. Private source-ledger provenance, real organization identities, private database identifiers and the transformation mappings are not released. Reproductions should record both the Hugging Face dataset commit and the Git commit.

## References

[1] C. M. Anderson, V. W. Fang, J. R. Moon Jr., and J. E. Shipman. Accounting for Cryptocurrencies. Journal of Accounting Research, 64(1):45–79, 2026.

[2] T. Patwardhan, R. Dias, E. Proehl, et al. GDPval: Evaluating AI Model Performance on Real-World Economically Valuable Tasks. arXiv:2510.04374, 2025.

[3] J. Jiang, C. Yang, A. Cui, et al. FinMaster: A Holistic Benchmark for Mastering Full-Pipeline Financial Workflows with LLMs. arXiv:2505.13533, 2025.

[4] R. Wang, J. Liu, W. Zhao, S. Li, and D. Zhang. AuditBench: A Benchmark for Large Language Models in Financial Statement Auditing. AI4Research and SEAS Workshops at AAAI 2025, CCIS vol. 2533, Springer, 2025. Preprint arXiv:2506.17282.

[5] H. Dong, P. Zhang, Y. Gao, et al. Finch: Benchmarking Finance & Accounting across Spreadsheet-Centric Enterprise Workflows. arXiv:2512.13168; Findings of ACL 2026.

[6] S. Tumpati, D. Agarwal, A. Kedia, et al. FinBalance: A Multi-Document Accounting Reconciliation Benchmark. arXiv:2606.15949, 2026.

[7] J. Benchek, A. Bennett, J. Kern, et al. APEX-Accounting. arXiv:2607.27189, 2026.

## Appendix A Evaluation Configuration

**Settings common to all 12 pipelines.** A maximum output of 32K tokens per turn; 1 exposed tool type, a bash command runner, callable repeatedly across up to 12 agent turns, with a structured-answer call after the agent loop where required; a 60-second timeout per command, with tool output truncated at 16,000 characters of stdout and 4,000 of stderr; the identical system contract and rendered task prompt; reasoning enabled at the highest setting each API exposes; 3 independent attempts per task. No pipeline configures top-p, top-k, a context-window limit, or a stop sequence.

**Execution environment.** Each attempt receives an empty temporary working directory, with HOME and TMPDIR pointed at it and API keys scrubbed from the bash subprocess environment. No reference answer, provenance record or evaluation artifact is copied into it, and evaluation loads the reference answer separately, after candidate inference. This is workspace isolation rather than an operating-system security sandbox: there is no chroot, no container boundary and no filesystem denylist, and network access is not disabled. Provider API mechanics, request timeouts and reasoning settings are provider-specific, so the benchmark contract and the exposed tool type are what the 12 pipelines share.

1 judge configuration, GPT-5.5 at the *high* reasoning setting, served every reported evaluation. It is recorded identically in all 4,248 evaluation artifacts, so 1 judging contract applies to every candidate model, and it is not among the 12 models in Table 3.

### Gate definitions and calibration

6 of the 7 required gates are deterministic and independent of the judge. 5 are computed from the candidate answer itself: parseable output, a balanced journal entry, the required journal structure, quantity consistency, and wallet custody correctness. The 6th verifies that the task's reference entry is gradeable, which is a property of the task rather than of the answer. The 7th, material accounting correctness, is computed from the frozen criterion verdicts for the criteria each task family designates as material, so the gate and the judge's accounting semantics cannot disagree. Being derived from those verdicts, it is satisfied automatically whenever every criterion passes. It applies at the attempt level across the whole score range, separating a materially wrong entry from an incomplete entry.

The deterministic gates are what make the Pass@3 conjunction bind. Of the 1,917 attempts that earn a perfect rubric score, 92 still fail a required gate. Those attempts are not necessarily wrong about the accounting: an entry can name the right accounts and carry the right asset quantity while departing from the recorded journal structure, or while writing an account name the reference does not use. The gates hold the reported metric to the recorded entry rather than to an equivalent rearrangement of it.

Calibration used deliberately corrupted reference entries rather than candidate answers. Replacing every amount with 999.99 leaves the entry balanced, because both sides are equally wrong, and clears every check except the material gate. It scores 0.85 on all 37 transfer, 12 fee, 27 income/expense and 16 swap tasks, and 0.90 on all 21 intercompany tasks, whose amount criterion weighs 10%. The smallest material criterion weighs 10–15%, so 1 material failure still lands at 0.85 or above, which is where the weighted score alone reads as nearly correct. A companion check runs the converse over the whole set: on all 118 reference entries the gates fire on no correct answer.

### Entry normalization

The deterministic checker also computes normalized complete-entry equality: every reference line pairs with 1 predicted line, with no missing and no extra lines, compared on account identity, Debit/Credit side, base-currency amount and currency. Account identity is read from the earliest populated of the ledger account name, account, account name and account id fields, with runs of whitespace collapsed and case folded. Lines whose amount is exactly decimal zero are dropped before comparison, and the remaining amounts are compared as canonical decimals. The rubric's amount criterion is a separate judgment, in which amounts are evaluated against the reference entry as part of criterion grading. This equality is an internal diagnostic and is not among the reported benchmark metrics.

| Model | Reasoning |
| --- | --- |
| Gemini 3.7 Flash | HIGH |
| Fable 5 | xhigh |
| GPT-5.6 Terra | xhigh |
| GPT-5.6 Sol | xhigh |
| Kimi K3 | max |
| Claude Sonnet 5 | xhigh |
| Grok 4.5 | high |
| Muse Spark 1.2 | xhigh |
| Grok 4.6 | high |
| Claude Opus 5 | xhigh |
| deepseek-v4-flash-0731 | max |
| GPT-5.6 Luna | xhigh |

*Table 3. Per-model run configuration.* ***Reasoning*** *is the value passed to the model's own reasoning parameter, which the APIs name differently: reasoning_effort, output_config.effort alongside adaptive thinking, or thinking_level.*

## Appendix B Additional Benchmark Statistics

CAB contains 55 inflows and 63 outflows. Derived entity relationships comprise 54 external, 22 intercompany, and 42 same-entity tasks. The complete chart of accounts is shown with every task. Transaction evidence spans created lots (17 tasks), relieved lots (21), tax-lot evidence of either kind (38), related same-hash records (60), and the prior-receipt history for the recurring transaction pattern (all 118).

### Failure-label semantics

Failure signals are deterministic, non-exclusive, and named in accounting terms. An incomplete entry is a required posting absent from the answer; an unsupported posting is an answer line with no counterpart in the recorded entry; a misposted account is a paired line whose account differs. A Debit/Credit criterion failure is a failure of the applicable Debit/Credit or intercompany balance-side criterion. That criterion compares the full set of account-and-side pairs for nonzero lines, so an account substitution, a missing line, or an extra line can raise it even without a side reversal. A wrong deciding account is a failure of the task's primary-account criterion, and an incorrect intercompany treatment is a treatment-criterion failure on a task whose rubric family is intercompany. 1 error can raise several signals, so counts describe diagnostic incidence, not a mutually exclusive causal taxonomy.

Some signals nest inside others: every wrong deciding account is also a misposted account, and every incorrect intercompany treatment is also an incorrect treatment. The leading pair is equal at 2,328 because both fire on precisely the same trajectories, not by coincidence of totals: a line whose account does not match is a misposted account, and it leaves the reference line it should have paired with unmatched.

*118 tasks; 12 candidate model pipelines; 3 attempts per task.*